\documentclass[a4paper, 10 pt, conference]{ieeeconf}  

\IEEEoverridecommandlockouts                           

\usepackage{float}
\usepackage{graphicx}
\usepackage{subfigure}
\usepackage{lipsum}
\usepackage{stfloats}
\usepackage{multicol}
\usepackage{multirow}
\usepackage{amsmath,bm}
\usepackage{etoolbox}
\usepackage{url}
\usepackage{array}

\usepackage[caption=false,font=footnotesize]{subfig}

\usepackage{bm}

\usepackage{textcomp}

\usepackage{colortbl}

\title{\LARGE \bf
Importance-Aware Semantic Segmentation with Efficient Pyramidal Context Network for Navigational Assistant Systems
}

\author{Kaite Xiang, Kaiwei Wang$^{*}$ and Kailun Yang
\thanks{This work has been partially funded through the project ``Research on Vision Sensor Technology Fusing Multidimensional Parameters'' (111303-I21805) by Hangzhou SurImage Technology Co., Ltd and supported by Hangzhou KrVision Technology Co., Ltd (krvision.cn).}
\thanks{Kaite Xiang, Kaiwei Wang and Kailun Yang are with State Key Laboratory of Modern Optical Instrumentation, Zhejiang University, Hangzhou, China {\tt\{katexiang, wangkaiwei, elnino\}@zju.edu.cn}}
}

\begin{document}

\maketitle
\thispagestyle{empty}
\pagestyle{empty}

\begin{abstract}

Semantic Segmentation (SS) is a task to assign semantic label to each pixel of the images, which is of immense significance for autonomous vehicles, robotics and assisted navigation of vulnerable road users. It is obvious that in different application scenarios, different objects possess hierarchical importance and safety-relevance, but conventional loss functions like cross entropy have not taken the different levels of importance of diverse traffic elements into consideration. To address this dilemma, we leverage and re-design an importance-aware loss function, throwing insightful hints on how importance of semantics are assigned for real-world applications. To customize semantic segmentation networks for different navigational tasks, we extend ERF-PSPNet, a real-time segmenter designed for wearable device aiding visually impaired pedestrians, and propose BiERF-PSPNet, which can yield high-quality segmentation maps with finer spatial details exceptionally suitable for autonomous vehicles. A comprehensive variety of experiments with these efficient pyramidal context networks on CamVid and Cityscapes datasets demonstrates the effectiveness of our proposal to support diverse navigational assistant systems.

\end{abstract}

\section{INTRODUCTION}
Semantic Segmentation (SS) is a task to assign semantic labels to each pixel of the images, which is of crucial significance for autonomous vehicles, robotics and navigation assistance systems for vulnerable road users like visually impaired pedestrians, where safety is critical~\cite{yang2019can}.

In recent years, with the development of deep learning, SS has come into the stage based on deep convolutional neural networks (CNNs) since the milestone created by Fully Convolutional Networks (FCN)~\cite{long2015fully}. The performance of FCN is surpassed by subsequent PSPNet~\cite{zhao2017pyramid} and DeepLab~\cite{chen2018deeplab}, which can perform semantic segmentation with high accuracies and huge numbers of parameters. Inevitably, the complex calculation keeps SS from being put into practice for real-time applications in devices with limited computation resources. In previous works, we propose ERF-PSPNet~\cite{yang2018unifying}\cite{yang2018semantic}, a real-time SS network especially designed for navigation assistance systems supporting the visually impaired, which largely sacrifices the resolution and accuracy of edges extraction, resulting in coarse segmentation maps. However, applications like autonomous vehicles and driving assistance require high-resolution semantic maps and highly accurate road boundary segmentation. To address this problem, we extend ERF-PSPNet with novel efficient pyramidal context network design by proposing BiERF-PSPNet, which can yield high-quality semantic segmentation maps with finer spatial details, while maintaining real-time inference.

\begin{figure}[t] 
  \centering 
  \subfigure[Image]{ 
    \includegraphics[width=0.42\linewidth]{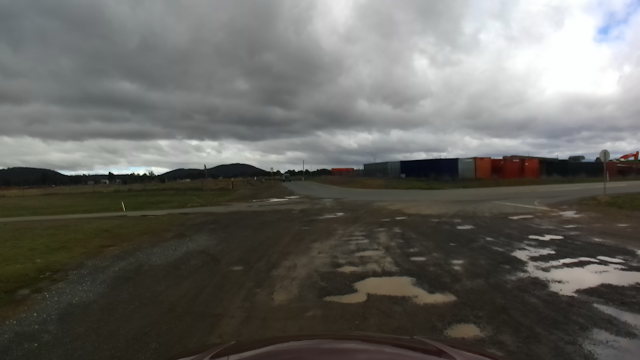} 
  } 
  \subfigure[Ground truth]{ 
    \includegraphics[width=0.42\linewidth]{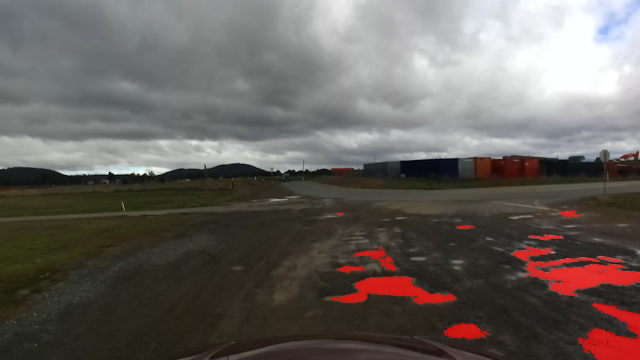} 
  } 
    \subfigure[Cross-entropy Loss]{ 
    \includegraphics[width=0.42\linewidth]{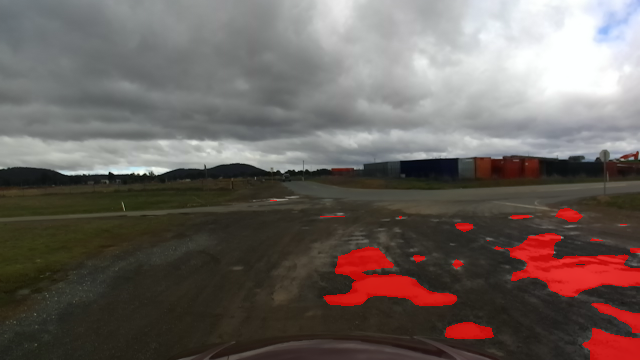} 
  } 
    \subfigure[Focal Loss]{ 
    \includegraphics[width=0.42\linewidth]{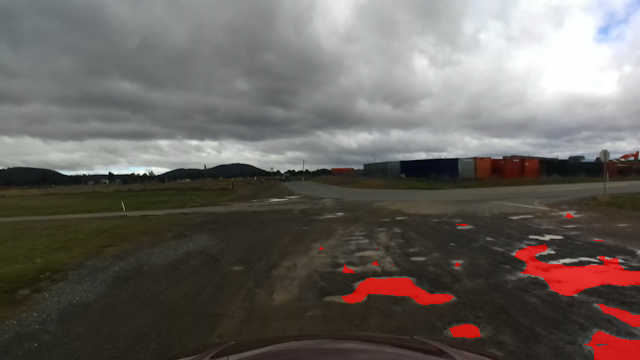} 
  } 
  \caption{Effect of different loss functions: (c) Output of ERF-PSPNet trained with cross-entropy loss, (d) Output of ERF-PSPNet trained with focal loss. It could observed that the model trained with cross-entropy loss has a higher recall rate and lower precision than cross focal loss.} 
\end{figure}
\begin{table}[t]
\caption{FOCAL LOSS V.S CROSS-ENTROPY LOSS}
\begin{center}
\begin{tabular}{|c|c|c|c|}
\hline
Loss & Recall rate & Precision & IoU\\
\hline
Focal loss & 0.80 & \textbf{0.85} & \textbf{0.70}\\
\hline
Cross-entropy loss & \textbf{0.94} & 0.63 & 0.60\\
\hline
\end{tabular}
\end{center}
\vskip-4ex
\end{table}

In addition to the architecture design, we made key observation that loss function is the key element in the training procedure and greatly influences the SS output. Traditional cross-entropy loss function is broadly used for SS, but the loss function only pays attention to the frequencies of objects by applying weights for different classes with different frequencies. For another thing, the focal-loss~\cite{lin2017focal} that is designed for difficulty-aware object detection task has also been utilized in existing researches to train SS networks. Evidently, the model trained with different loss functions will have a vast difference between recall rate and precision. Before our project, we conduct an experiment on a water puddle segmentation dataset~\cite{han2018single}. We find that the model trained with focal loss possesses a higher precision while the model trained with cross-entropy loss has a higher recall rate, as shown in Fig. 1 and Table~\uppercase\expandafter{\romannumeral1}. We argue that in some safety-critical application scenarios of autonomous vehicles, recall rate plays a more important role than precision for traffic objects like cars, buses, and pedestrians, as we need to focus on the detection of them, which need to be detected with high recall rate. In other words, it is preferred to detect it wrongly rather than miss it, because these traffic objects will be dangerous if the algorithm misses them and predict them as safe roadways. In addition, hierarchical importance should be emphasized for different objects for autonomous vehicles. Taking it for granted, roadways and sidewalks are more important than sky and buildings, while cars and pedestrians are even more important and safety-critical than those flat classes.

Therefore, existing methodologies like focal loss and cross-entropy loss are not ideally suitable for SS associated with autonomous driving system. As explained above, an autonomous driving system needs to focus on some important objects for driving rather than segment all classes with the same level of importance. In this paper, inspired by~\cite{chen2018importance}, we adapt and re-design an importance-aware loss function (IAL), and perform a comprehensive set of experiments to prove its effectiveness and wide applicability.

The contributions of the paper lie in four key aspects:
\begin{itemize}

\item We adapt a real-time SS architecture named ERF-PSPNet, and extend the model into a bilateral architecture BiERF-PSPNet to recover better spatial details.
\item We adapt and re-design an importance-aware loss function, and improve its stability and reliability.
\item A series of experiments are conducted on two autonomous driving benchmark, \textit{i.e.}, CamVid~\cite{brostow2008segmentation} and Cityscapes~\cite{cordts2016cityscapes}, which demonstrate the structure of the refined model can recover better spatial information and the effectiveness of the adapted IAL.
\item We perform a systematic analysis of the experiment results, throwing insightful hints on how importance is assigned for real-world SS frameworks. Our implementations and codes are available at: \url{https://github.com/Katexiang/E
RF-PSPNET}

\end{itemize}

\section{Related work}

\subsection{Semantic Segmentation Neural Networks}

Since the milestone created by FCN, SS has gain tremendous advances based on CNNs. The ConvNets first transfer known classification networks into SS by making them fully convolutional. Immediately following the success, UNet~\cite{ronneberger2015u}, DeepLab~\cite{chen2018deeplab} and many other SS Networks were proposed. Many of them have achieved state-of-the-art performance on different benchmarks of SS task. Their normal procedure involves encoding more spatial information or enlarging the receptive filed at the expenses of huge operations and multiple parameters. Therefore, they normally perform inference at a low speed so that they can not be applied for real-time application like autonomous vehicles.

In order to put the SS networks into practice, many light-weighted real-time SS networks were proposed. ENet~\cite{paszke2016enet} is one of the first networks in pursuit of real-time inference, which is modified from ResNet structure~\cite{he2016deep} to perform SS with much fewer parameters. ERFNet~\cite{romera2018erfnet}~\cite{romera2019bridging} and our previous ERF-PSPNet~\cite{yang2018unifying}\cite{yang2018semantic} utilize residual factorized module to reduce parameters and keep fine performance.

At the same time, some SS networks with multi-path structure were put forward to refine the spatial details of the output. ICNet~\cite{zhao2018icnet} is one of the pioneer with multi-path structure, which uses multiple-size input image at shallow layers to get spatial information, while inputting small image to deep layers to extract semantic information. BiSeNet~\cite{yu2018bisenet} works in a different way, which divides the network structure into two paths, one for spatial information to refine output, and another for excavating context information. ContextNet~\cite{poudel2018contextnet} combines a deep network branch at low resolution capturing global context efficiently with a shallow branch focusing on high-resolution segmentation details to reach competitive performance. 
\subsection{Somewhat-Aware Method for training CNNS}
With the development of deep learning, many training methods are advanced to solve somewhat-aware problems like difficult-aware~\cite{li2017not} and attribute-aware~\cite{sulistiyo2018attribute} SS. Li et al.~\cite{li2017not} considered that different pixels own different ranks of difficulty and propose a difficulty-aware network to pay attention to more difficult pixels. At the same time, focal loss~\cite{lin2017focal} acts as a loss function to cope with the detection of difficult objects. Inspired by attention mechanism, Chen et al.~\cite{chen2016attention} designed a SS network to emphasize objects with different scales.  Bulo et al.~\cite{bulo2017loss} introduced a novel loss max-pooling concept for handing imbalanced training data distributions. Following it, Importance-Aware-Loss (IAL) created by Chen et al.~\cite{chen2018importance} was leveraged to distinguish important pixels from normal pixels. But IAL is unstable, sometimes the effect is remarkable, and sometimes it is unserviceable. To alleviate the shortcomings, in this paper we re-design and adapt to a more stable IAL, and prove the effectiveness.

As is shown above, among the various existing notions of SS networks, they will be coming into use in the near future. Besides, the somewhat-aware method can be exploited to cope with certain application problems in somewhat-bias tasks like importance, scale, difficulty and so on. Based on these observations, this paper aims to cope with the problem that different objects own different levels of importance in autonomous driving systems. 

\section{Methodology}

In this section, we firstly detail the modified version of Importance-Aware-Loss (IAL), and then illustrate our real-time SS networks, ERF-PSPNet and its extended version, BiERF-PSPNet. 

\subsection{Importance-aware Loss Function} 
IAL proposed by Bi et al.~\cite{chen2018importance} is a modified version of entropy-cross loss function in practice. 
 Making a brief introduction to traditional entropy-cross loss function \textbf{I} which is defined by:
$$
\textbf{I} = -\sum_{i=1}^{H} \sum_{j=1}^{W} \textbf{q$_{i,j}$}\cdot log(  \textbf{p$_{i,j}$})  \eqno{(1)}
$$
where \textbf{q$_{i,j}$} and \textbf{p$_{i,j}$} are the one-hot encoding label and output at i-th row and j-th column, both of which have the shape of (1,C) (C: the number of classes). When training the model, \textbf{I} needs to be divided by \textbf{H} and \textbf{W}, which represents height and width of the image respectively.
However, the model trained with the loss function may segment certain pixels into classes that occupies most pixels of the image. Therefore, weight $\omega_{i,j}$ can be employed in the loss function 
to enable the model to pay attention to the rare classes. It is defined by:
$$
\omega_{i,j}= {1\over ln(\textbf{a}+\textbf{f$_{i,j}$})}  \eqno{(2)}
$$
where \textbf{a} is a hyper-parameter avoiding divided by zero, in this paper, we set it to 1.02. And \textbf{f$_{i,j}$} is the pixel sum of the class at i-th row and j-th column divided by the number of  input image's pixels. Therefore, \textbf{I} can be modified into
$$
\textbf{I} = -\sum_{i=1}^{H} \sum_{j=1}^{W} \omega_{i,j}\cdot \textbf{q$_{i,j}$}\cdot log(  \textbf{p$_{i,j}$})  \eqno{(3)}
$$

In order to enable models to focus on certain important objects, we need to exert dynamic weight for loss function. Taking CamVid as an example, the dataset has 11 classes, \textit{i.e.}, sky, building, pole, road, sidewalk, tree, sign, fence, car, pedestrian and bicyclist, which will be detailed in Section \uppercase\expandafter{\romannumeral4}. First, we categorize the classes into three importance groups as a hierarchical structure like Fig. 2. For autonomous driving systems, the traffic objects like cars, pedestrians are the most important, while the road or sidewalks are less important, the sky and buildings away from the passable area are the least important. 
\begin{figure}[t] 
  \centering 
  \subfigure[]{ 
    \includegraphics[height=3cm]{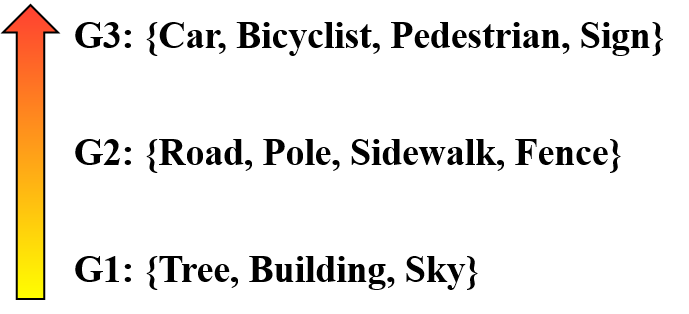} 
  } 
  \subfigure[\textbf{M$_{1}$}]{ 
    \includegraphics[height=3cm]{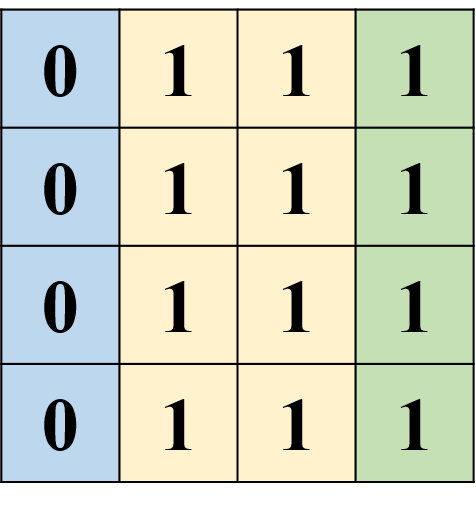} 
  } 
    \subfigure[\textbf{M$_{2}$}]{ 
    \includegraphics[height=3cm]{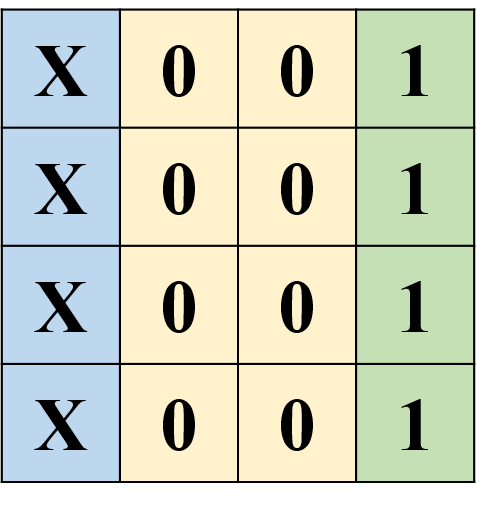} 
  } 

  \caption{(a) shows the rankings of importance of CamVid classes, G3 is the most important group. (b) and (c) are the importance matrices, (b) illustrates the \textbf{M$_{1}$}, (c) illustrates the \textbf{M$_{2}$}, where the blue area belongs to G1, the yellow area belongs to G2, and the green area belongs to G3. } 
\end{figure}
Then we divide the \textbf{I} into three parts, \textit{i.e.}, \textbf{I$_{1}$}, \textbf{I$_{2}$} and \textbf{I$_{3}$}, each of which stands for the certain part of the cross-entropy loss belonging to certain group. Following the rationale, here comes to importance matrix, \textbf{M$_{i}$}. If we have three ranks of importance, we should construct two importance matrices, \textit{i.e.}, \textbf{M$_{1}$} and \textbf{M$_{2}$} as shown in Fig. 2. Taking \textbf{M$_{2}$} as an example, for three ranks of importance category, the most important classes are assigned to 1 at both matrices, while the classes of middle rank are signed to 1 at \textbf{M$_{1}$} and 0 at \textbf{M$_{2}$}, and the least important classes are assigned to 0 at \textbf{M$_{1}$} and \textbf{X} at \textbf{M$_{2}$} (\textbf{X} is a number either 0 or 1). They are the key elements to dynamically assign importance weights for the loss function.

Afterwards, we need to utilize the matrices to construct the dynamic importance weights. The dynamic weight  of a group $f_{t}$ is defined as
$$
  {\sum\sum{[(\textbf{M$_{t,i,j}$}+\lambda)^{0.5}\cdot (\textbf{p$_{c,i,j}$}-\textbf{M$_{t,i,j}$})\cdot(\textbf{M$_{t,i,j}$}\ne \textbf{X})]^2}\over N_{t}} \eqno{(4)}
$$
where $f_{t}$ (
 t can be chosen as 2 or 3 in three-rank importance system as the G1's importance weight is 0) is the dynamic importance weight; $\lambda$ is a tuning parameter set to 0.5 in order to take the lower-importance category into consideration and avoid ignoring them when calculating the dynamic importance weight. \textbf{M$_{t,i,j}$} is the value of the importance matrix, while \textbf{p$_{c,i,j}$} is the ground-truth channel value of the output at i-th row and j-th column; \textbf{p$_{c,i,j}$} is the key element of the weight pushing the loss function focusing on important category. And the value of ($\textbf{M$_{t,i,j}$}\ne \textbf{X}$) is 0 if the value of the matrix is \textbf{X} else the value is 1. $N_{t}$ is a normalization factor, which is the pixel sum of the full image when t is 2 and the pixel sum of G2 and G3 when t is 3.

At present, the proposed loss function can be defined by
$$
\textbf{IAL} = \textbf{I$_{1}$}+(f_{1}+\alpha)\cdot\textbf{I$_{2}$}+(f_{2}+\alpha)\cdot(f_{3}+\alpha)\cdot\textbf{I$_{3}$}  \eqno{(5)}
$$
where \textbf{IAL} is the ultimate loss function, $\alpha$ is a tuning parameter being set to 1 in our experiment.

\subsection{Architecture}
In view of the trade-off between efficiency and accuracy, we select an efficient pyramidal context network, \textit{i.e.}, ERF-PSPNet~\cite{yang2018unifying} as our base net. As is shown in Fig. 3(a), the model follows a typical encoder-decoder architecture. The model is a rational combination of efficient residual factorized network (ERFNet) and pyramid scene parsing network (PSPNet). The encoder originates from ERFNet, which utilizes a sequential architecture to produce down-sampled feature maps. The encoder first utilize the ``down-sampler'' block as detailed in~\cite{paszke2016enet} to down-sample the feature map quickly in order to reduce computation costs. The highlight of the encoder is ``Non-bottleneck-1D'' as detailed in~\cite{romera2018erfnet} enabling an efficient utilization of minimized amount of residual layers to extract effective feature maps and achieve high efficiency. Following the pyramid pooling module modified from PSPNet, the decoder is designed to harvest contextual information among feature maps of varied sizes and attain larger receptive field. After that, the feature maps are bilinearly interpolated and cascaded to form the final feature representation. Following the concatenation layer, we append a convolution layer to re-weight the feature representation. In the end, we append a 1$\times$1 kernel classification convolution layer, bilinear interpolation layer and softmax layer to output the final result.
\begin{figure*}[t] 
  \centering 
  \subfigure[ERF-PSPNet]{ 
    \includegraphics[height=3.5cm]{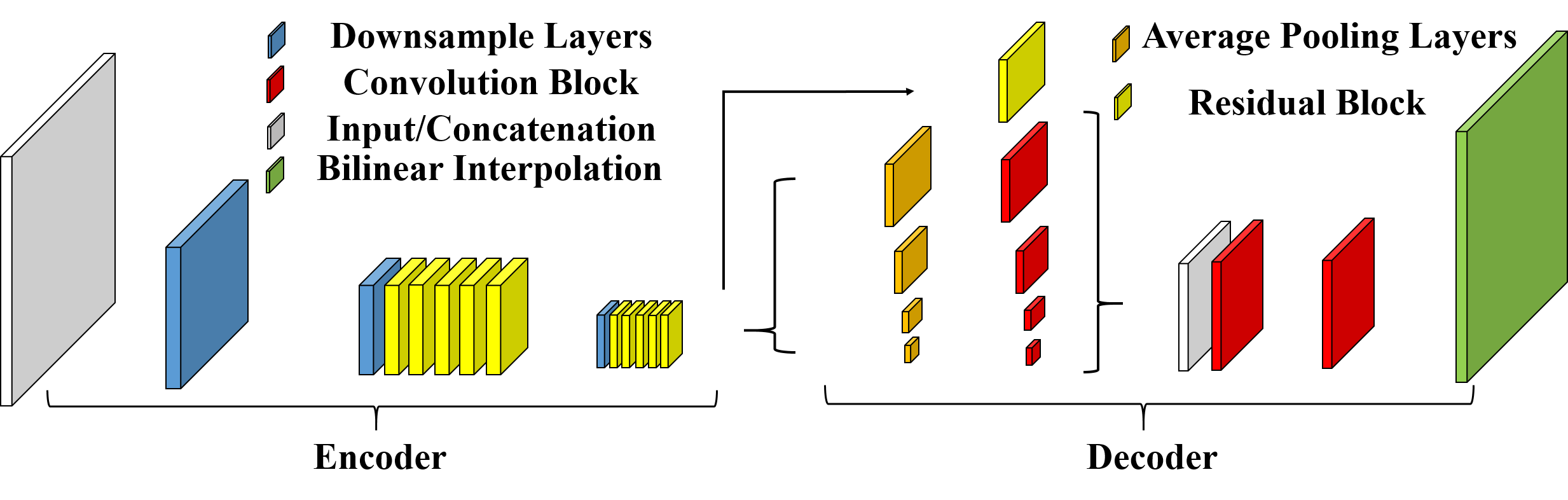} 
  } 
  \subfigure[BiERF-PSPNet]{ 
    \includegraphics[height=3.2cm]{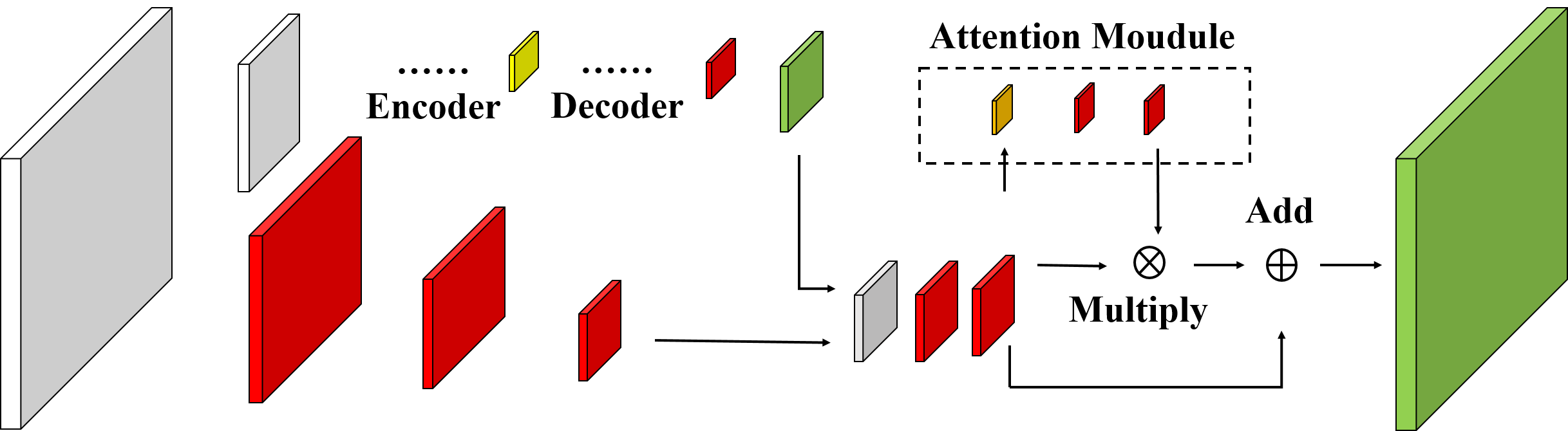} 
  } 
  \caption{The architecture of our semantic segmentation networks: (a)~ERF-PSPNet and (b)~BiERF-PSPNet.} 
\end{figure*}

The SS network is especially designed for assisted navigation of the visually impaired. Therefore, the segmentation at the semantics boundaries (\textit{e.g.}, road boundaries) remain coarse, because the decoder attains better contextual information at the sacrifice of texture and spatial information, which are less desired by visually impaired pedestrains, but these details are important for autonomous driving. Inspired by bilateral BiseNet and ContextNet, we advance ERF-PSPNet by proposing an important variant of the SS network with a different style. Making a brief introduction of BiSeNet, it is composed of spatial path and context path. The context path is a feature extractor to attain deep semantic information at low computation cost. In contrast, the spatial path is a concatenation of several convolution layers as shallow layers to extract spatial information fed with a high-resolution image. Therefore, BiSeNet's output is not only of high precision but also possesses fine textures. Therefore, another efficient pyramidal context network, \textit{i.e.}, modified ERF-PSPNet is proposed concerning the textures of the output, named BiERF-PSPNet as shown in Fig. 3(b). Different from BiSeNet, our proposed model possesses two input images, the smaller one of which serves as the context information source going through deep neural network to attain semantic information, and the larger one of which is used for extracting spatial information to refine the results. In the end, feature maps of the two paths are fused by an attention module as shown in Fig. 3(b). In our experiments, \textbf{H$_{1}$} and \textbf{W$_{1}$} are 1024 and 2048 respectively. \textbf{H} and \textbf{W} are 360 and 720 respectively.

\section{Experiments}
\begin{table*}[t]
\caption{CROSS-ENTROPY LOSS V.S IAL on Camvid by ERF-PSPNet (\%). \protect\\G3,G2 and G1 are the groups of the classes. G3 is the most important group and G1 is the least important group.}
\begin{center}
\begin{tabular}{|c|c|c|c|c|c|c|c|c|c|c|c|c|}
\hline 
\multicolumn{2}{|c|}{Group } & \multicolumn{4}{c|}{G3} &\multicolumn{4}{c|}{G2} &\multicolumn{3}{c|}{G1}\\
\hline 
\multicolumn{2}{|c|}{Class} & Sign & Car & Pedestrian & Bicyclist& Pole & Road & Sidewalk & Fence& Sky & Building & Tree\\
\hline
\multirow{3}*{Cross-Entropy}
& Precision & 33.0& 86.2& 47.8& 69.4& 35.9& 93.8& 83.3& 39.9& 94.5& 85.6& 76.7\\
\cline{2-13}&
 Recall rate & 34.0& 79.9& 63.7& 47.2 & 42.6& 97.1& 81.9& 25.3& 94.2& 82.3&80.5\\
\cline{2-13}& IoU & 20.1&70.9& 37.6& 39.1& 24.2&91.2&70.3& 18.3& 89.3&72.3&64.7\\
\hline
\multirow{3}*{IAL}& Precision & 31.0& 83.1& 39.3& 65.6& 31.3& \textbf{94.1}&81.1 &34.1& \textbf{94.8}&\textbf{86.2} &76.6\\
\cline{2-13}& Recall rate & \textbf{47.0}& \textbf{83.4}& \textbf{69.1}& 42.7 & 40.6& 96.4& \textbf{83.3}& \textbf{27.3}& 93.0& 79.5& 78.7\\
\cline{2-13}& IoU & \textbf{23.0}&\textbf{71.3}& 33.5& 34.9& 21.5&90.9&69.7& 17.8& 88.5&70.6&63.4 \\
\hline
\end{tabular}
\end{center}
\vskip-4ex
\end{table*}
\begin{table*}[t]
\caption{CROSS-ENTROPY LOSS V.S IAL on Cityscapes of G3 by BiERF-PSPNet (\%)}
\begin{center}
\begin{tabular}{|c|c|c|c|c|c|c|c|c|c|}
\hline 
\multicolumn{2}{|c|}{Class} & Traffic Light& Sign& Rider& Truck&Bus&Train&Motorcycle&Bicycle\\
\hline
\multirow{3}*{Cross-Entropy}
 & Precision & 72.8& 81.8& 61.3& 73.1& 69.6& 64.5& 46.5& 77.7\\
\cline{2-10}&
 Recall Rate &62.0& 73.4& 50.3& 62.2 & 72.2& 22.4& 34.3& 77.0\\
\cline{2-10}&
 IoU & 50.4&63.1& 38.2& 50.6& 54.9&19.9&24.6& 63.0\\
\hline
\multirow{3}*{IAL} & Precision & 63.5& 73.6 & \textbf{61.4}& 72.6& \textbf{70.0}&\textbf{78.7} &\textbf{51.6}& 76.2\\
\cline{2-10}&
 Recall rate & \textbf{68.9}& \textbf{78.4}& 47.6& 61.5 & \textbf{76.7}& \textbf{46.0}& 34.1& \textbf{79.7}\\
\cline{2-10}&
 IoU &49.4&61.2&36.6& 49.9& \textbf{57.7}&\textbf{40.8}&\textbf{25.8}&\textbf{63.8} \\
\hline
\end{tabular}
\end{center}
\vskip-4ex
\end{table*}
\begin{table*}[t]
\caption{CROSS-ENTROPY LOSS V.S IAL on Cityscapes of G2 and G1 by BiERF-PSPNet (\%)}
\begin{center}
\begin{tabular}{|c|c|c|c|c|c|c|c|c|c|c|c|c|}
\hline 
\multicolumn{2}{|c|}{Group } & \multicolumn{5}{c|}{G2} &\multicolumn{6}{c|}{G1}\\
\hline 
\multicolumn{2}{|c|}{Class} & Car& Sidewalk& Fence& Pole& Pedestrian& Road& Building& Wall& Vegetation& Terrain& Sky\\
\hline
\multirow{3}*{Cross-Entropy} & Precision & 94.1&84.1&65.4&70.0&71.4&98.7&92.9&63.6&94.3&73.9&94.2\\
\cline{2-13} & Recall rate &95.4&88.5&47.2&65.4&87.7&98.1&94.0&41.5&94.7&68.3&97.6\\
\cline{2-13} & IoU &90.0&75.8&37.7&51.0&64.9&96.8&87.7&33.6&89.6&55.0&92.1\\
\hline
\multirow{3}*{IAL} & Precision &93.0&81.1&63.7&65.1&69.4&\textbf{98.9}&\textbf{93.4}&\textbf{64.4}&\textbf{94.4}&70.0&\textbf{94.9}\\
\cline{2-13} & Recall rate &\textbf{96.1}&\textbf{89.6}&\textbf{47.5}&\textbf{67.0}&\textbf{88.6}&97.5&93.0&\textbf{45.8}&93.6&65.4&97.1\\
\cline{2-13} & IoU &89.5&74.2&37.4&49.3&63.7&96.5&87.3&\textbf{36.5}&88.7&51.0&\textbf{92.3}\\
\hline
\end{tabular}
\end{center}
\vskip-4ex
\end{table*}

\begin{table*}[h]
\caption{CROSS-ENTROPY LOSS V.S IAL on Camvid and Cityscapes of mean Groups(\%)}
\begin{center}
\begin{tabular}{|c|c|c|c|c|c|c|c|c|c|}
\hline 
\multicolumn{2}{|c|}{Dataset} & \multicolumn{4}{c|}{Camvid by ERF-PSPNet} & \multicolumn{4}{c|}{Cityscapes by BiERF-PSPNet}\\
\hline 
\multicolumn{2}{|c|}{Group} & 3 & 2 &1&Mean & 3 & 2 &1&Mean\\
\hline
\multirow{3}*{Cross-Entropy}
 & Precision & 59.1& 63.2& 85.6&67.8& 68.4& 77.0& 86.3&76.3\\
\cline{2-10}&
  Recall rate &56.2& 61.7&85.7&66.2&56.7&76.9&82.4&70.1\\
\cline{2-10}&
IoU & 41.9&51.0&75.4&54.3&45.6&63.9&75.8&60.0\\
\hline
\multirow{3}*{IAL}
 & Precision & 54.8&60.2 &\textbf{85.9}&65.2&68.4&74.5 &86.0&75.6\\
\cline{2-10}& Recall rate &\textbf{60.6}& \textbf{61.9}& 83.7&\textbf{67.4}&\textbf{61.6}& \textbf{77.8}& 82.17&\textbf{72.3}\\
\cline{2-10}&IoU &40.7&50.0&74.2&53.2&\textbf{48.2}&62.8&75.4&\textbf{60.6}\\
\hline
\end{tabular}
\end{center}
\vskip-4ex
\end{table*}

In order to verify our model and IAL, we conduct a series of experiments on dataset, \textit{i.e.}, CamVid and Cityscapes.
\subsection{Datasets}
\textit{$CamVid$} : The dataset is a street scene dataset from a driving vehicle's perspective containing 701 images of 720 $\times$ 960 and involving 11 semantic categories, 367 of which belong to its training set and 233 of which belong to the validation set.

\textit{$Cityscapes$} : The dataset is a street scene dataset from the perspective of an intelligent car, which contains 2975 fine annotated images for training and another 500 images for validation. The resolution of the images is 1024 $\times$ 2048, we select 19 pre-defined classes for training and validation.
\subsection{Implementation Details}
We use Tensorflow and a NVDIA GeForce GTX 1080Ti GPU for training and validation. Due to the limited memory we set batch size to 8 when training ERF-PSPNet and 5 when training BiERF-PSPNet. ERF-PSPNet's mIoU can reach 59.7 at a highly efficient resolution of 360 $\times$ 720, but it can be enhanced to 64.5 when adopting data augmentations. In order to facilitate fair comparison, we abandon data augmentations for our experiments inspite of the effect for advancing performance and robustness of the models. The models are trained for 300 epochs (for both datasets) with Adam optimization algorithm. The initial learning rate is 0.001 divided by 10 every 100 epochs. For the sake of combatting overfitting, we use the L2 weight regularization with decay of 0.0002.
\subsection{Quantitative Results}
\begin{figure*}[t] 
  \centering 
  \subfigure[Image]{ 
    \includegraphics[width=0.231\linewidth]{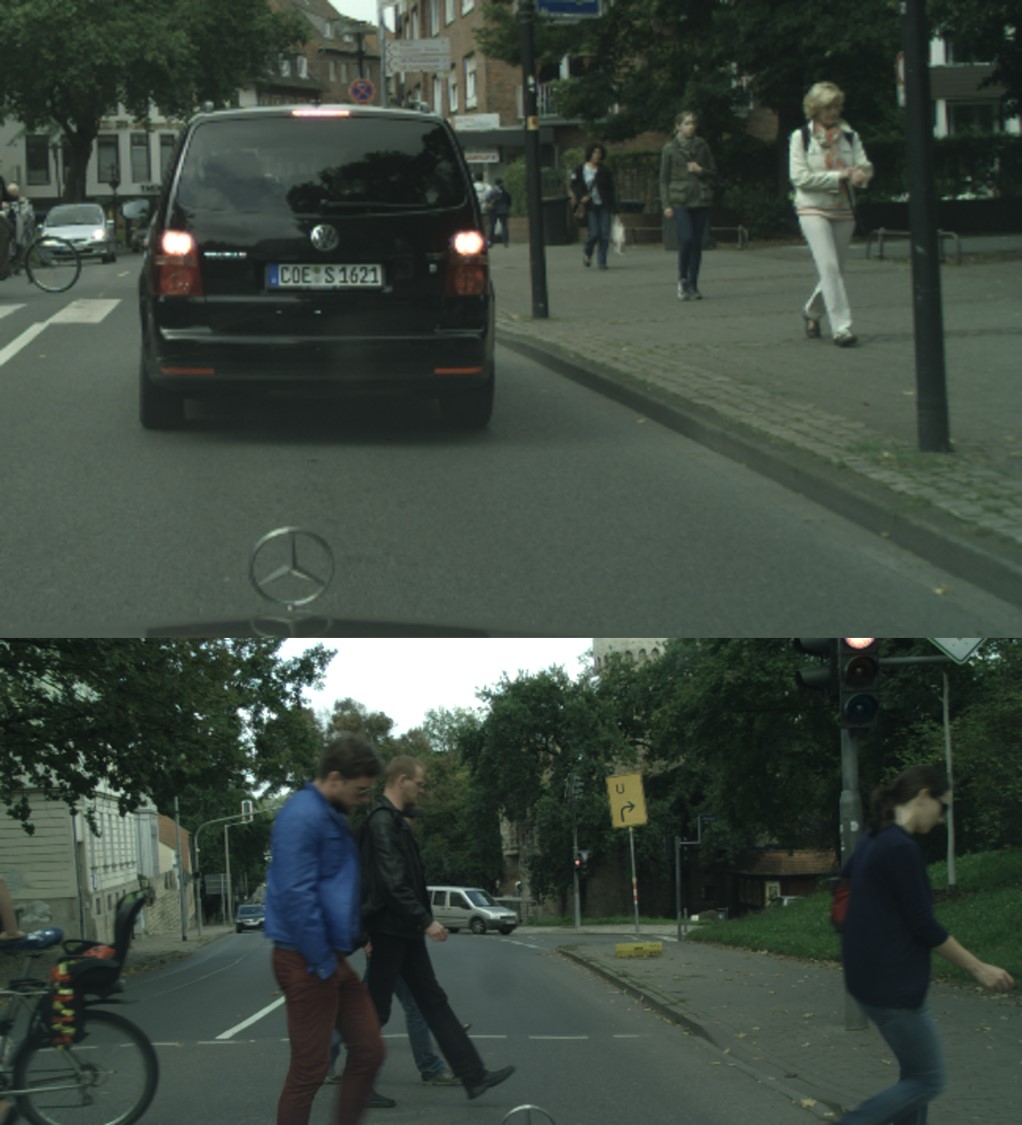} } 
  \subfigure[Label]{ 
    \includegraphics[width=0.231\linewidth]{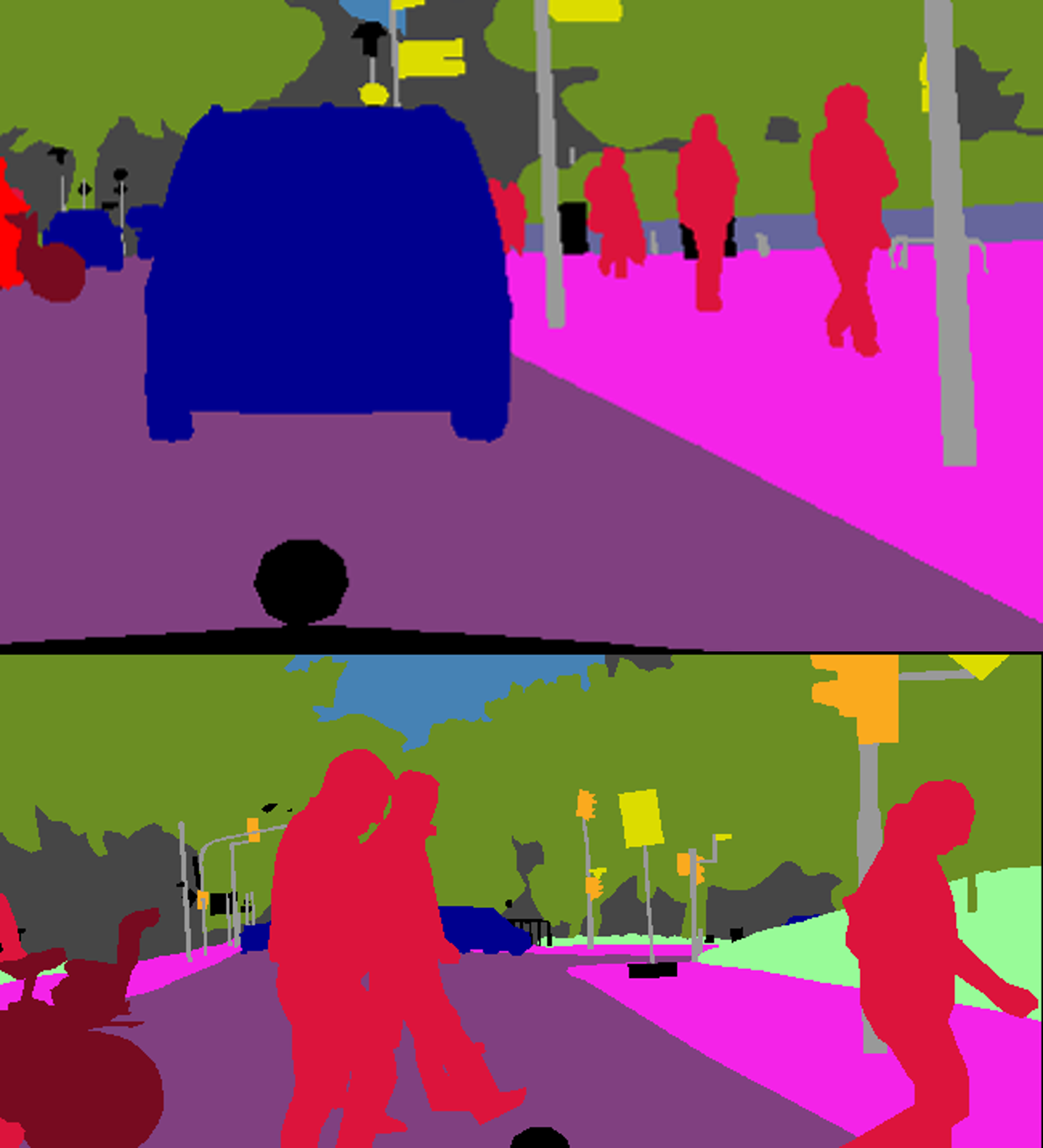} } 
    \subfigure[Result of ERF-PSPNet (coarse)]{ 
    \includegraphics[width=0.231\linewidth]{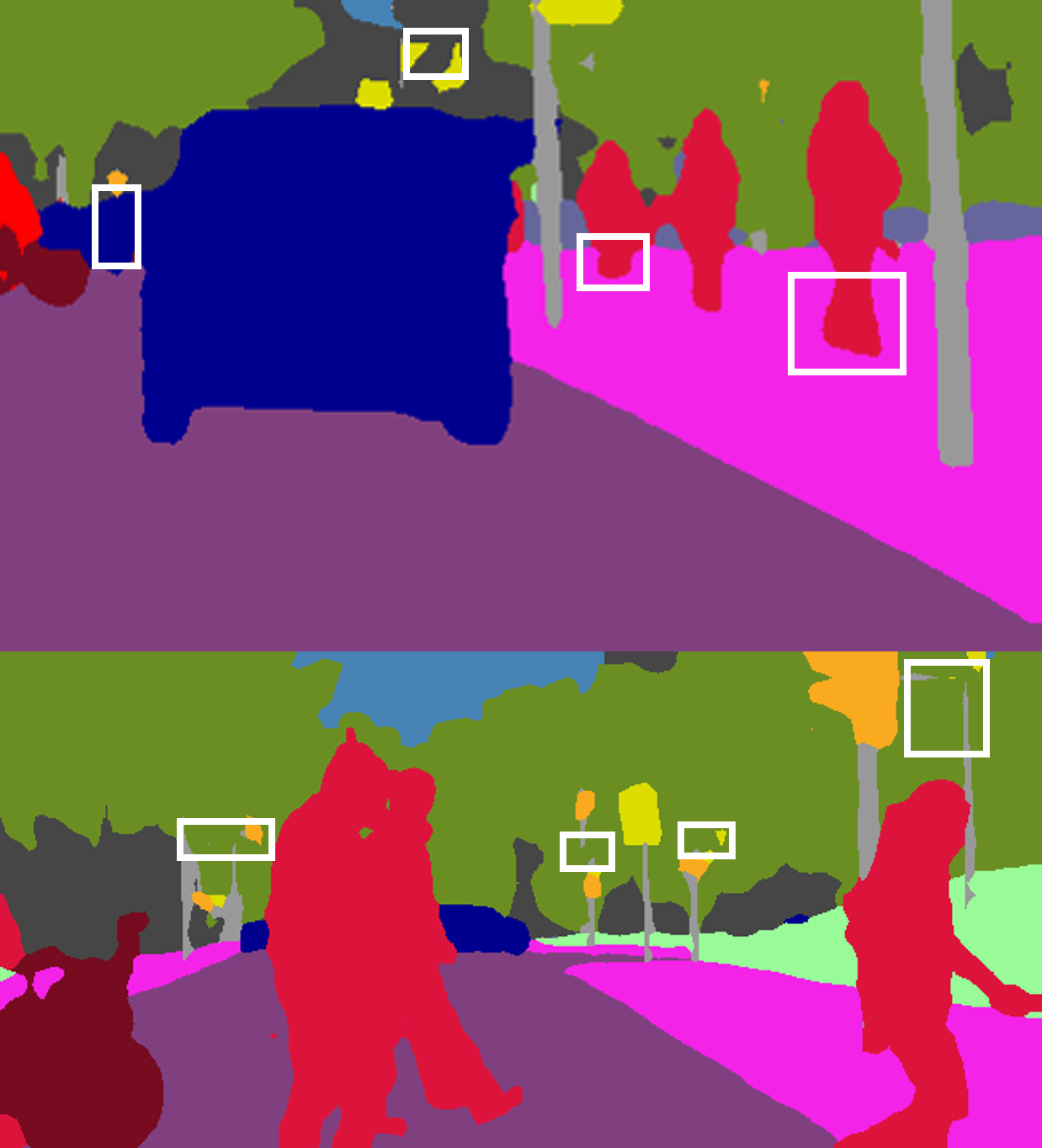} } 
    \subfigure[Result of BiERF-PSPNet (fine)]{ 
    \includegraphics[width=0.231\linewidth]{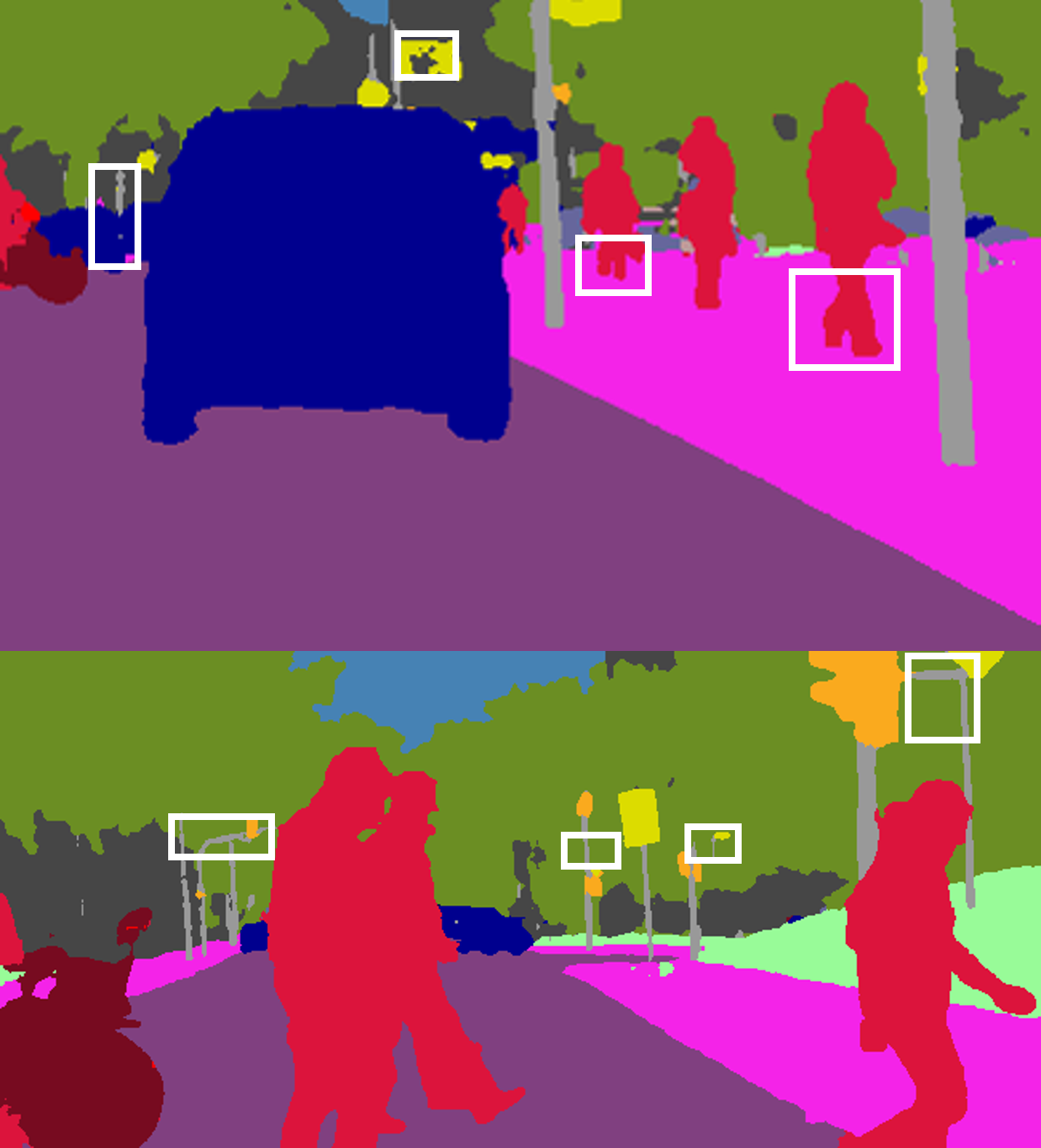} } 
  \caption{The result comparison between ERF-PSPNet and BiERF-PSPNet. } 
\end{figure*}
\begin{figure*}[t] 
  \centering 
  \subfigure[Image]{ 
    \includegraphics[width=0.231\linewidth]{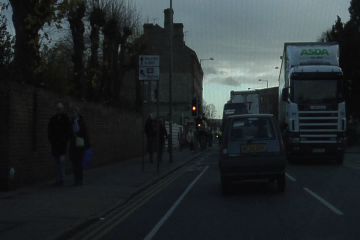} 
  } 
  \subfigure[Label]{ 
    \includegraphics[width=0.231\linewidth]{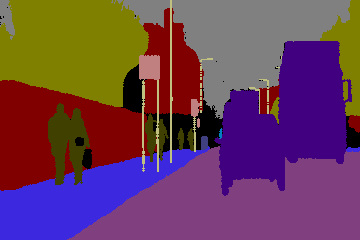} 
  } 
    \subfigure[Cross-entropy Loss]{ 
    \includegraphics[width=0.231\linewidth]{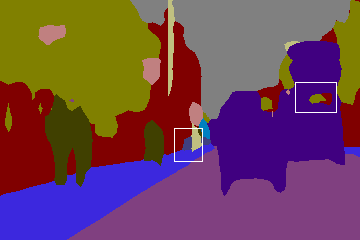} 
  } 
    \subfigure[IAL]{ 
    \includegraphics[width=0.231\linewidth]{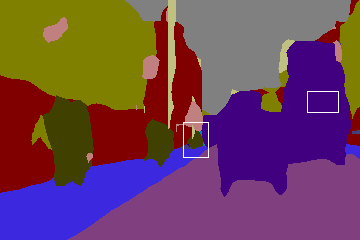} 
  } 
  \caption{The result comparison between Cross-entropy loss and IAL in CamVid.} 
\end{figure*}
\begin{figure*}[t] 
  \centering 
  \subfigure[Image]{ 
    \includegraphics[width=0.231\linewidth]{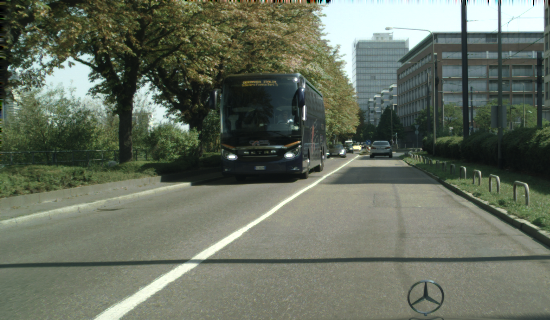} 
  } 
  \subfigure[Label]{ 
    \includegraphics[width=0.231\linewidth]{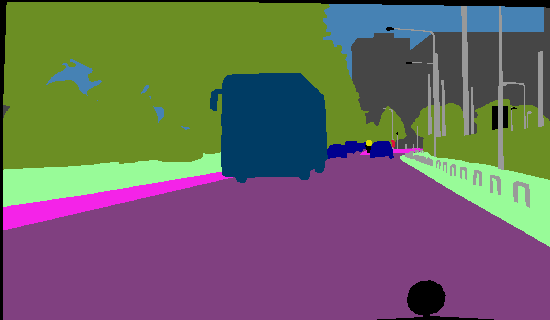} 
  } 
    \subfigure[Cross-entropy Loss]{ 
    \includegraphics[width=0.231\linewidth]{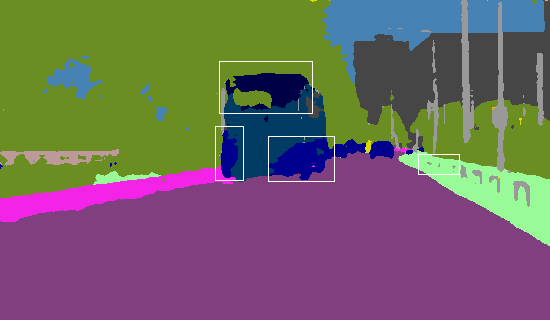} 
  } 
    \subfigure[IAL]{ 
    \includegraphics[width=0.231\linewidth]{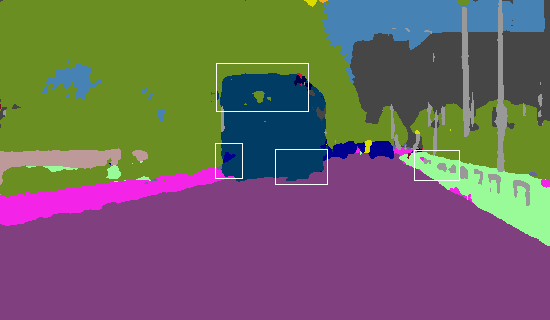} 
  } 
  \caption{The result comparison between Cross-entropy loss and IAL in Cityscapes.} 
\end{figure*}
\begin{figure}[h] 
  \centering 
\includegraphics[width=0.8\linewidth]{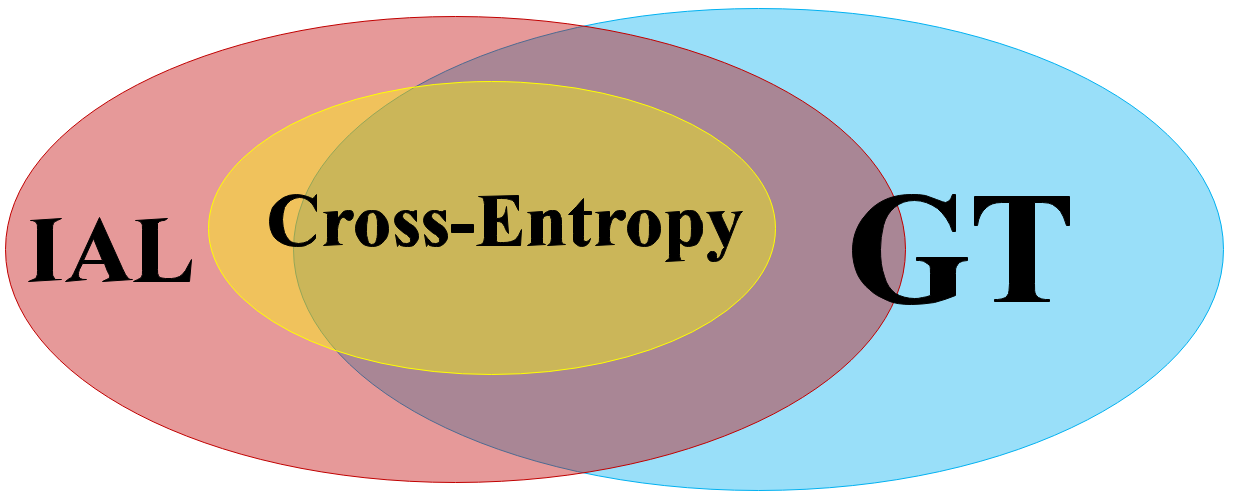} 
  \caption{Graphical illustration of the effect of IAL for G3. The blue area is ground truth, the yellow are is the output of model trained by cross-entropy loss, and the red area is the result of the model trained by IAL.} 
\end{figure}

We conduct three groups of experiments, one of which is conducted on CamVid, and others are conducted on Cityscapes.

For CamVid, we select the ranks of importance as depicted in Fig. 2(a). From the results of Table \uppercase\expandafter{\romannumeral2} and Table \uppercase\expandafter{\romannumeral5}, we observe that the recall rates of G3 have been advanced especially the classes like sign, car and pedestrian, and the mean recall rate has been improved by around 1 point. In view of the different categories between CamVid and Cityscapes, some categories' precision and recall rate are quite high for Cityscapes like pedestrians, cars and road. On the other hand, the extra categories which do not belong to CamVid are more important than them. Therefore, some categories need to be attributed into different importance rank or else it may lead to detrimental effect for training. For Cityscapes, we regard traffic light, sign, rider, truck, bus, train, motorcycle, bicycle as the most important classes, car, sidewalk, fence, pole, pedestrian as the second important classes, and road, building, wall, vegetation, terrain, sky as the least important classes. We conduct a series of experiments by using ERF-PSPNet and BiERF-PSPNet, both of them demonstrate the effectiveness of IAL. Taking BiERF-PSPNet's results as an example, the results are filled in Table \uppercase\expandafter{\romannumeral3} to Table \uppercase\expandafter{\romannumeral5}. From the results, what we can learn is that the IAL elevates the important classes' recall rates dramatically with few negative effect on the precision. 

The speed of ERF-PSPNet is 74.1fps when inputting a 360 $\times$ 720 image on a GTX 1080Ti GPU and BiERF-PSPNet is 42.1fps. Their mIoU are 59.7 and 60.7 on Cityscapes validation set, respectively. In other words, the BiERF-PSPNet refine the spatial information by making a sacrifice for inference time, while still keeping above real-time inference. 

However, surprisingly, a by-product of IAL, attracts our interests, which is the promotion of the G1's precision on both datasets as displayed in Table \uppercase\expandafter{\romannumeral2}, Table \uppercase\expandafter{\romannumeral4} and Table \uppercase\expandafter{\romannumeral5}. In other words, when categorizing the classes into three importance parts, although the original purpose is to advance the recall rate of G3, the precision of G1 has been advanced and even has a slight improvement on mIoU by accident, which is of practical significance for autonomous vehicles and other navigational assistant systems. We have emphasized the importance of recall rate for autonomous vehicles in Section \uppercase\expandafter{\romannumeral1}, but precision is another key point. In comparison, regarding navigation assistance for the visually impaired, we may underline the segmentation of sidewalks, which should be segmented with high precision, because the system must guarantee the visually impaired people navigate on safe sidewalks, in case the road is detected as sidewalks which will be dangerous for them.

\subsection{Qualitative Analysis}
The effect of BiERF-PSPNet can be shown in Fig. 4. We find that the edge of objects is more accurate and refined because of the spatial path of the BiERF-PSPNet, especially at the edges of the pedestrians, riders, and some small and slender objects like telegraph poles, which are very important for safety-critical autonomous driving, as it is required to perceive pedestrians and poles at long distances, in order to take fast decisions in response to environmental events.

We find that IAL is of great effect in Fig. 5 and Fig. 6. In Fig. 5 from CamVid, we find the output of the cross-entropy loss ignores some pedestrians at further places and the segmentation of the truck is fragmented bounded by white boxes. But in the IAL's output, the fragmentation of the truck is refined, while the smaller pedestrians can be detected. Fig. 6 shows a representative example from Cityscapes, where it is obvious that the IAL is highly effective successfully to segment the bus and detect most of the poles while the cross-entropy loss's model segments part of the bus into car, part of the bus into truck. Moreover, it misses some poles as well. Therefore, we find our advanced IAL is not only effective in CamVid but also practical in challenging large-scale Cityscapes as well.

\section{Conclusions and Future Work}
SS is promising for many tasks especially navigational assistant systems like autonomous driving. As different objects possess different ranks of importance, the SS network for autonomous driving should be addressed by a different method, \textit{i.e.}, our revised IAL can yet be regarded as a powerful technique. This paper re-design IAL so that the loss function can make the training model focus on important classes and advance the classes' recall rate which will impose dynamic weights adaptively. In addition, we adapt ERF-PSPNet into BiERF-PSPNet for the sake of a finer spatial result, while maintaining above real-time inference.

As a saying goes, every rose has its thorn. The revised IAL can advance the recall rate of the important classes and precision of the least important classes without damaging the performance of the model, but it decreases the precision of the important classes which may category other objects into the important classes. Therefore, we summarize the performance of IAL as the Venn Diagram shown in Fig. 7. For G3, IAL has a higher recall rate and lower precision than cross-entropy loss, while keeping competitive mIoU. In other words, training with IAL yields models which would rather segment the important objects wrongly, than miss them for safety considerations. In the future, we aim to further optimize IAL and attempt to utilize new decision rules~\cite{chan2019application} to let model to improve certain categories’ recall rate and certain categories' precision.


\bibliographystyle{IEEEtran}
\bibliography{root}

\end{document}